\documentclass[10pt,twocolumn,letterpaper]{article}

\usepackage[pagenumbers]{cvpr} 

\usepackage{graphicx}
\usepackage{amsmath} 
\usepackage{amssymb}
\usepackage{booktabs}

\usepackage[pagebackref,breaklinks,colorlinks]{hyperref}

\usepackage[capitalize]{cleveref}
\crefname{section}{Sec.}{Secs.}
\Crefname{section}{Section}{Sections}
\Crefname{table}{Table}{Tables}
\crefname{table}{Tab.}{Tabs.}

\usepackage{amsmath,amsfonts,bm}

\def\eqref#1{equation~\ref{#1}}

\def\1{\bm{1}}

\DeclareMathAlphabet{\mathsfit}{\encodingdefault}{\sfdefault}{m}{sl}
\SetMathAlphabet{\mathsfit}{bold}{\encodingdefault}{\sfdefault}{bx}{n}

\DeclareMathOperator*{\argmax}{arg\,max}

\usepackage{booktabs}
\usepackage{makecell}
\usepackage{xcolor}
\usepackage{lipsum}
\usepackage{comment}
\usepackage{multirow}
\usepackage{wrapfig}
\usepackage{algorithm}
\usepackage{algorithmic}
\usepackage{subfigure}

\usepackage{graphicx}
\usepackage{color}

\usepackage[english]{babel}
\usepackage{amsthm}
\definecolor{darkgreen}{rgb}{0,0.7,0}

\newcommand{\green}[1]{{\color{darkgreen}{#1}}}
\usepackage{colortbl}
\usepackage{xcolor}
\definecolor{mygray}{gray}{.6}
\newcommand{\gray}[1]{{\color{mygray}{#1}}}

\newcommand{\D}{\mathcal{D}}

\renewcommand{\L}{\mathcal{L}}

\newcommand{\M}{\mathcal{M}}

\def\eg{\emph{e.g.}} 
\def\ie{\emph{i.e.}}

\def\wrt{{w.r.t.\ }}

\def\cvprPaperID{5157} 
\def\confName{CVPR}
\def\confYear{2022}

\begin{document}

\title{DyRep: Bootstrapping Training with Dynamic Re-parameterization}

\author{%
	Tao Huang${}^{1,2}$ \quad Shan You$^{1}$\thanks{Correspondence to: Shan You $<$\texttt{youshan@sensetime.com}$>$.}  \quad Bohan Zhang$^3$\\
	Yuxuan Du$^2$ \quad Fei Wang$^4$ \quad Chen Qian$^1$ \quad Chang Xu$^2$\\
	\normalsize $^1$SenseTime Research\quad
	\normalsize $^2$School of Computer Science, Faculty of Engineering, The University of Sydney\\
	\normalsize $^3$ College of Information and Computer Sciences, University of Massachusetts Amherst\\
	\normalsize $^4$University of Science and Technology of China \\
}

\maketitle

\begin{abstract}
Structural re-parameterization (Rep) methods achieve noticeable improvements on simple VGG-style networks. Despite the prevalence, current Rep methods simply re-parameterize all operations into an augmented network, including those that rarely contribute to the model's performance. As such, the price to pay is an expensive computational overhead to manipulate these unnecessary behaviors. To eliminate the above caveats, we aim to bootstrap the training with minimal cost by devising a dynamic re-parameterization (DyRep) method, which encodes Rep technique into the training process that dynamically evolves the network structures. Concretely, our proposal adaptively finds the operations which contribute most to the loss in the network, and applies Rep to enhance their representational capacity. Besides, to suppress the noisy and redundant operations introduced by Rep, we devise a de-parameterization technique for a more compact re-parameterization. With this regard, DyRep is more efficient than Rep since it smoothly evolves the given network instead of constructing an over-parameterized network. Experimental results demonstrate our effectiveness, e.g., DyRep  improves the accuracy of  ResNet-18 by $2.04\%$ on ImageNet and reduces $22\%$ runtime over the baseline. Code is available at: \href{https://github.com/hunto/DyRep}{https://github.com/hunto/DyRep}.

\end{abstract}


\section{Introduction} \label{sec:intro}
The advent of automatic feature engineering fuels deep convolution neural networks (CNNs) to reach the remarkable success in a plethora of computer vision tasks, such as image classification~\cite{he2016deep, xie2017aggregated, howard2017mobilenets, zheng2021weakly}, object detection~\cite{ren2016faster,lin2017feature,fang2021instances}, and semantic segmentation~\cite{zhao2017pyramid,he2017mask}. In the path of pursuing better performance than that of early prototypes such as VGG~\cite{simonyan2014very} and ResNet~\cite{he2016deep}, current deep learning models ~\cite{xie2017aggregated,hu2018squeeze,li2019selective} generally are embodied with billions of parameters and paramount well-tailored architectures and operations (\eg, channel-wise attention in SENet~\cite{hu2018squeeze} and branch-concatenation in Inception~\cite{szegedy2015going}). From this perspective, we may encounter a dilemma in the sense that a learning model with good performance should be heavy and computationally intensive, which is extremely hard to deploy and has a high inference time. To this end, a critical question is: \textit{how to enhance the ability of neural networks without incurring expensive computational overhead and high inference complexity?}

\begin{figure}[t]
    \centering
    \includegraphics[width=0.9\linewidth]{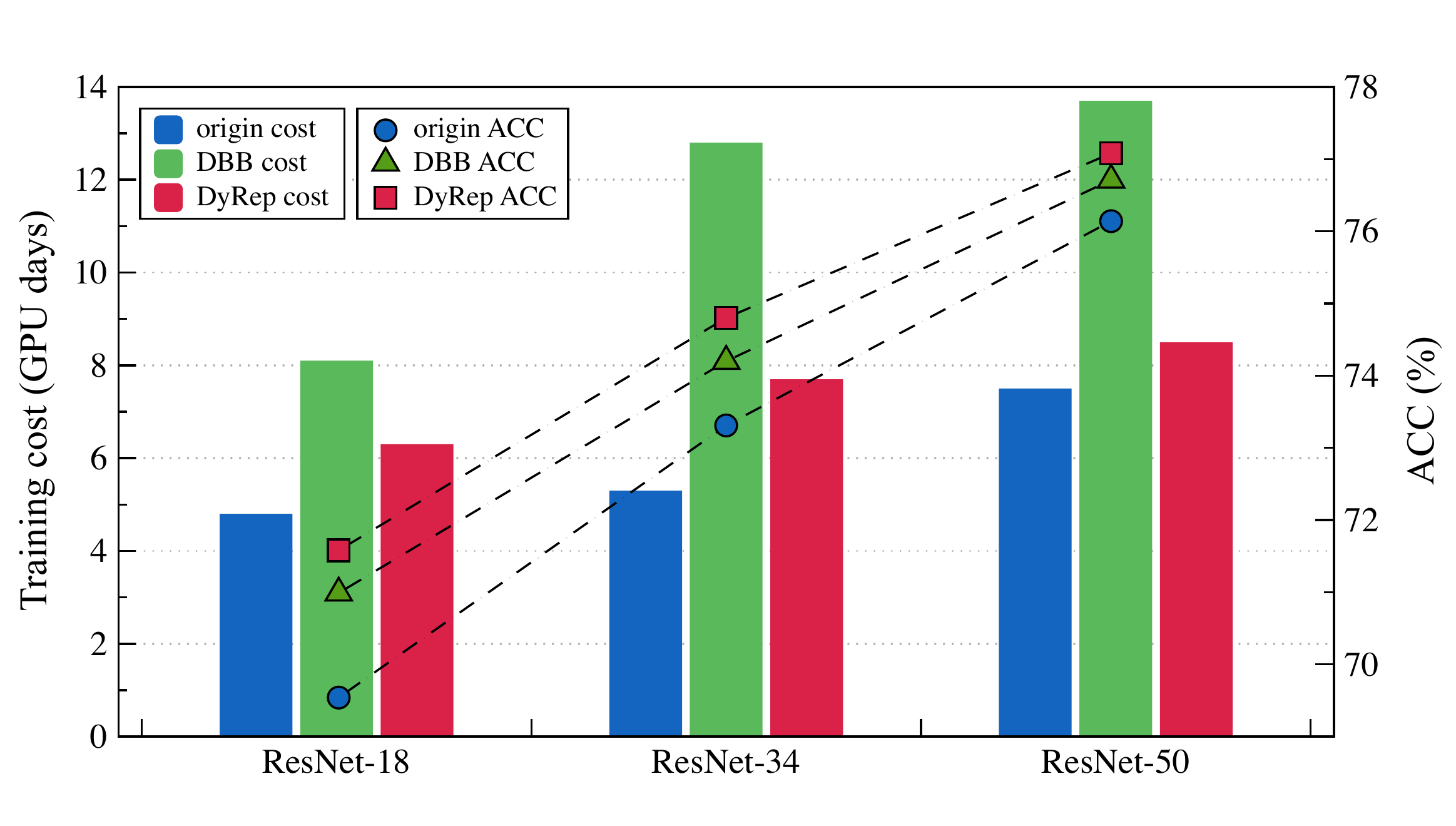}
    \vspace{-4mm}
    \caption{Accuracy and training cost of ResNet on ImageNet dataset using origin, DBB, and our DyRep models. Our DyRep obtains the highest accuracies yet has much smaller training cost compared to DBB.}
    \vspace{-4mm}
\end{figure}
\begin{figure*}[t]
    \centering
    \includegraphics[width=0.9\linewidth]{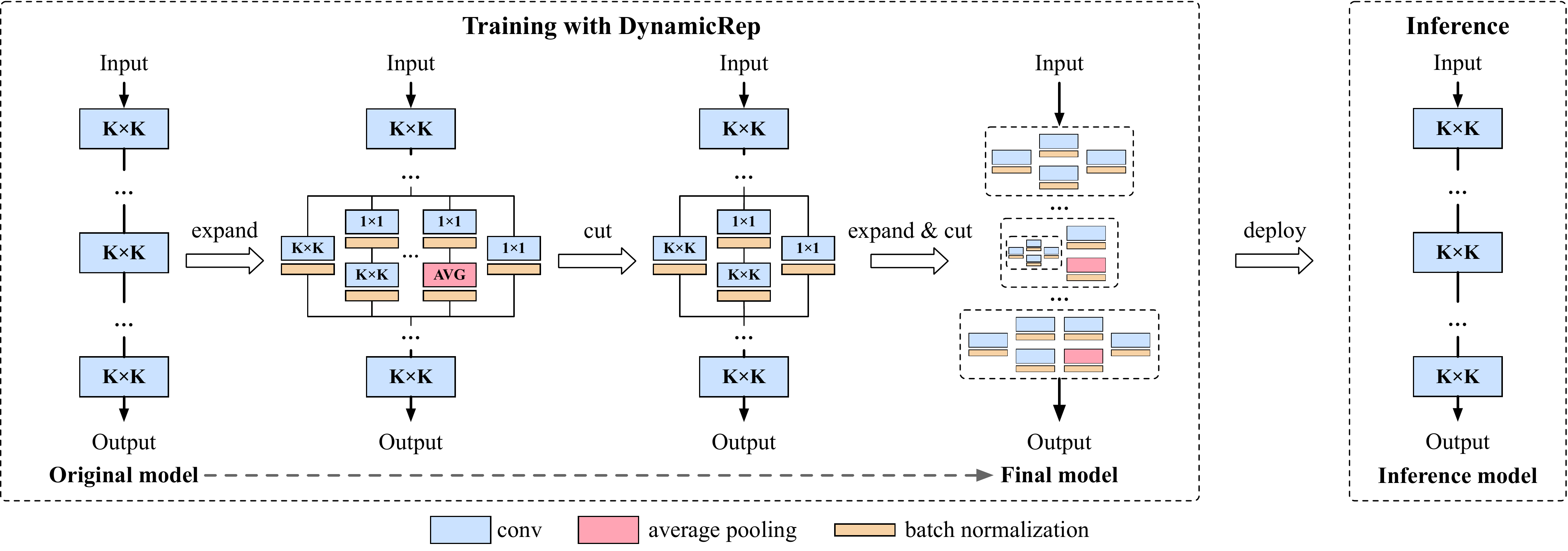}
    \vspace{-2mm}
    \caption{Overview of  Dynamic Re-parameterization (DyRep). Train (left panel): Starting from a simple model, DyRep dynamically adjusts the network structures in training by expanding operations to multi-branch blocks or cutting redundant branches. Inference (right panel):  The trained model is transformed to the original model for inference.}
    \label{fig:figure2}
\end{figure*}

Structural re-parameterization technique (Rep) and its variants \cite{ding2021repvgg,ding2021diverse,zhang2021repnas}, which construct an augmented model in training and transform it back to the original model in inference, have emerged as a leading strategy to address the above issue. Concretely, these methods enhance the representational capacities of models by expanding the original convolution operations with multiple branches in training, then fusing them into one convolution for efficient inference without accuracy drop. Representative examples include RepVGG~\cite{ding2021repvgg}  and DBB~\cite{ding2021diverse}. The former enhances VGG-style networks by expanding the $3\times 3$ \textit{Conv} to an accumulation of three branches (i.e., $3\times 3$ \textit{Conv}, $1\times 1$ \textit{Conv}, and \textit{residual connection}) in the training process and re-parameterizing it back to the original $3\times 3$ \textit{Conv} in the inference time. The latter improves CNNs by enriching the types of expanding branches  (i.e., introducing $6$ equivalent transformations of re-parameterization) and unifying them into a universal building block which applies to various CNNs such as ResNet~\cite{he2016deep} and MobileNet~\cite{howard2017mobilenets}. Nevertheless, a common caveat of current Rep and its variants is coarsely re-parameterizing all branches into an augmented network, where a large portion of them may seldom enhance the model’s performance. In other words, directly utilizing the same branches in all layers would lead to suboptimal structures. Furthermore, these redundant operations would 
result in an expensive or even unaffordable memory and computation cost, since the memory consumption increases linearly with the number of branches.

To conquer the aforementioned issues, we propose a novel re-parameterization method, dubbed as DyRep, to \textit{dynamically} evolve network structures during training and recover to the original network in inference, as illustrated in Figure~\ref{fig:figure2}. In particular, the key concept behind our proposal is adaptively seeking the operations with the biggest contributions to the performance (or loss as its surrogate) instead of pursuing a universal re-parameterization to all of them, which ensures both the efficacy and accuracy to augment the network. In DyRep, the operation with the biggest contribution amounts to the operation with the most significant \textit{saliency score}. As our first technical contribution, this measure is partially inspired by the gradient-based pruning methods, which utilize the gradients \wrt the loss to calculate the saliency scores of filters. 

Since the existing Rep methods are designed for transforming the model to a narrow one at the end of the training, there is no plug-and-play technique to expand one convolution to multiple branches while keeping the training stable. To achieve a training-aware Rep, we first extend the Rep technique in such a case, then propose to stabilize the training by initializing the additional branches with small scale factors in batch normalization (BN) layers. By doing so, the additional branches would start with minor importance, making trivial changes on the original weights, and thus obtaining a smooth structure evolution.

Our second key technical contribution is devising a de-parameterization method to unearth and discard the redundant operations that appeared in Rep. Since we initialize the BN layer in the newly-added branch with small scale factor, which can be treated as a relaxed gate to turn on or cut off one branch. That is, if a branch has a significant small scale value compared to other branches, it will make a minor contribution to the outputs. Therefore, we could discard it and absorb its weights to other branches for better efficiency. Specifically, if a branch has a zero scale factor, its operations would not affect the output.

Our main contributions are summarized as follows.
\begin{itemize}
    \item We propose DyRep, a dynamic re-parameterization method that applies to training, aiming to enhance the Rep performance with minimal overhead. By identifying the important operations dynamically during training, our proposal achieves  significant efficiency and performance improvement.
    \item Our DyRep is more friendly to downstream tasks such as object detection. Different from previous Rep and NAS methods that need to first train a network on image classification task, followed by transferring it to downstream tasks, DyRep can directly adapt the structures in downstream tasks. This property dramatically reduces the computational cost. 
    \item Extensive experiments on image classification and its downstream tasks demonstrate that DyRep outperforms other Rep methods in the measure of both the accuracy and the runtime cost.
\end{itemize}

\section{Related Work}
\subsection{Network Morphism}
Network morphism~\cite{wei2016network,wei2017modularized,dong2020towards} aims to morph one layer into multiple layers or discard multiple layers into one layer meanwhile preserving the function of the original network. These methods dynamically adjust computation workloads in different training stages, \ie, starting with a shallow network and gradually increasing its depth during training. However, since the growth of the network would change its outputs, additional training strategies (\eg, mimic learning) should be involved to minimize the reconstruction error between the new and original networks. As a result, various initialization methods, e.g., identity initialization~\cite{wei2016network}, random initialization~\cite{wen2020autogrow}, and initialization with partial training~\cite{istrate2018incremental}, have been proposed to warrant an effective training. In this paper, we achieve network morphism with negligible reconstruction error using Rep; hence the added operations could be randomly initialized without additional training steps, and the morphed network can be transformed back to the original network for efficient inference.

\subsection{Neural Architecture Search}
Neural architecture search (NAS) methods~\cite{zoph2016neural,liu2018darts,you2020greedynas,su2021prioritized,su2020locally} achieve remarkable performance improvements by automatic architecture designing. However, they are computationally expensive in training intermediate architectures. Although some one-shot NAS methods~\cite{liu2018darts,you2020greedynas} are proposed to reduce the runtime cost by regarding the whole search space as a supernet and training it once, it still suffers from a high memory consumption and an additional cost on supernet training. Recently,  RepNAS~\cite{zhang2021repnas} is proposed to search better Rep architectures by leveraging the differentiable NAS method~\cite{liu2018darts}. In this way,  RepNAS can directly transform the trained supernet into the final network in inference using Rep, without training the searched network again. However, RepNAS still suffers from the expensive computational overhead, as its training interacts with the whole search space (i.e., networks with all the Rep branches equipped). In this work, instead of utilizing NAS to pursue a fixed Rep structure, DyRep assumes that the optimal structures vary at different training phases (epochs) and aims to bootstrap the training with minimal cost. As such, DyRep uses Rep to dynamically evolves the network structures during training. Since our method starts training with the original networks, it would save much computation cost compared to RepNAS.

\begin{table*}
	\renewcommand\arraystretch{1.17}
	\setlength\tabcolsep{3mm}
	\centering
	\caption{Operation spaces of RepVGG, DBB, RepNAS, and DyRep. $K\times K$ denotes \textit{conv} operation with the kernel size $K\times K$, and $1\times1$-$K\times K$ denotes a branch stacking $1\times 1$ and $K\times K$ \textit{Conv} sequentially.}
	\vspace{-2mm}
	\label{tab:operation_space}
	\footnotesize
	\begin{tabular}{lcl}
		\Xhline{2\arrayrulewidth}
		Method & \#Branches & Branches\\
		\hline
		RepVGG~\cite{ding2021repvgg} & 3 & $K\times K$, $1\times1$, \textit{residual connection}\\
		DBB~\cite{ding2021diverse} & 4 & $K\times K$, $1\times1$-$K\times K$, $1\times1$-AVG, $1\times1$ \\
		RepNAS~\cite{zhang2021repnas} & 7 & $K\times K$, $1\times1$-$K\times K$, $1\times1$-AVG, $1\times1$, $1\times K$, $K\times1$, \textit{residual connection}\\
		DyRep (ours) & 7 & $K\times K$, $1\times1$-$K\times K$, $1\times1$-AVG, $1\times1$, $1\times K$, $K\times1$, \textit{residual connection}\\
		\Xhline{2\arrayrulewidth}
	\end{tabular}
	\vspace{-4mm}
\end{table*}

\section{Revisiting Structural Re-parameterization}
Let us first recap the mechanism of the vanilla structural re-parameterization (Rep) method \cite{ding2021repvgg,ding2021diverse}. The core ingredients of Rep are the equivalent transformations of operations. Concretely, these transformations do not only enhance the representational capacities of neural networks by introducing diverse branches in the training process, but can also be equivalently transformed to simpler operations, which promise a lighten neural networks in inference. These properties are of great importance in model mining, and enable to dramatically reduce the computational cost without losing accuracy. In the rest of this section, we revisit such transformation techniques used in Rep.

Rep engineers with operations that are widely integrated in networks, such as convolution (\textit{Conv}), average pooling and residual connection. For example, \textit{Conv} functions by transforming an input feature $\bm{I}\in\mathbb{R}^{C\times H\times W}$ to an output $\bm{O}$, \ie, 
\begin{equation}
    \bm{O} := o(\bm{I} ) =\bm{I} \circledast \bm{F} + \bm{b} \in\mathbb{R}^{D\times H'\times W'},
\end{equation}
where $C$, $H$, and $W$ refer to channels, height, and width of the input, respectively. $\bm{F}\in\mathbb{R}^{D\times C \times K \times K}$ and $\bm{b}\in\mathbb{R}^{D}$ are the parameters of the convolution operator $\circledast$. Note that $H'$ and $W'$ are determined by several factors such as kernel size, padding, stride, \textit{etc}. 

The linearity of the convolution operator $\circledast$ ensures the additivity. Specifically, for any two convolutions $o^{(1)}$ and $o^{(2)}$ with weights $\bm{F}^{(1)}$ and $\bm{F}^{(2)}$, if they follow the same configurations (\eg, the same $D$, $C$, and $K$), we have
\begin{equation} \label{eq:add}
    \bm{I} \circledast \bm{F}^{(1)} + \bm{I} \circledast \bm{F}^{(2)} = \bm{I} \circledast (\bm{F}^{(1)} + \bm{F}^{(2)}) .
\end{equation}
For ease of understanding, the derivation of Eq.(\ref{eq:add}) omits the term $\bm{b}$, but the above results still hold when $\bm{b}$ is considered. Supported by the additivity in  Eq.(\ref{eq:add}), an immediate observation is that two compatible \textit{Conv} operations can thus been merged into a single new \textit{Conv} operation $o^{(3)}$ with weights $\bm{F}^{(3)}=(\bm{F}^{(1)} + \bm{F}^{(2)})$. 

\begin{figure*}[t]
    \centering
    \includegraphics[width=1.0\linewidth]{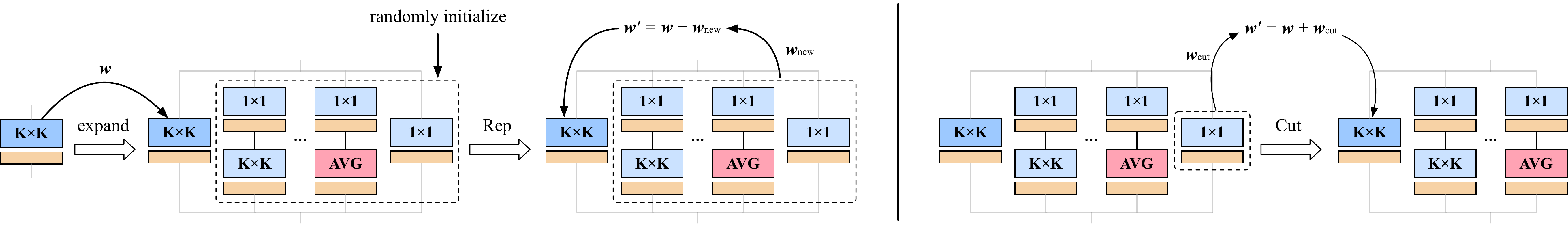}
    \vspace{-4mm}
    \caption{Illustration of  re-parameterization (left) and de-parameterization (right) methods used in DyRep. Re-parameterization: we first extend the original $K\times K$ \textit{Conv} using additional branches with randomly initialized weights, then leverage Rep to modify the weights in original convolution for a consistent output. De-parameterization: we remove the redundant operation by absorbing its weights into the original convolution.}
    \label{fig:figure3}
\end{figure*}

Note that the above additivity can be generalized to other operations once they can be transformed to a convolution operation. This evidences that  multi-branch operations, or equivalently a sequence of operations, can be transformed into a single convolution and thus possess the additivity. Without loss of clarity, we follow the convention in \cite{ding2021diverse} to refer a \textbf{branch} as an operation involved in the transformation. Some examples satisfying this rule are listed as follows. See the left panel in Figure~\ref{fig:figure3} for intuition.

$\bullet$ \textbf{A conv for sequential convolutions.} A sequence of \textit{Convs} $1\times 1\mbox{-}K\times K$ can be merged into one $K\times K$ \textit{Conv}. 

$\bullet$ \textbf{A conv for average pooling.} A $K\times K$ \textit{average pooling} is equivalent to a $K\times K$ \textit{Conv} with the same stride.

$\bullet$ \textbf{A conv for multi-scale convolutions.} $K_H\times K_W$ ($K_H\le K, K_W\le K$) convolutions (\eg, $1\times 1$ \textit{Conv} and $1\times K$ \textit{Conv}) can be transformed into a $K\times K$ \textit{Conv} via zero-padding on kernel weights. 

$\bullet$ \textbf{A conv for residual connections.} A residual connection can be viewed as a special $1\times 1$ \textit{Conv} with value $1$ everywhere, and thus can be transformed into a $K\times K$ \textit{Conv}.

By leveraging the above elementary transformations, one $K\times K$ \textit{Conv} can be augmented by adding more diverse branches to its output. For example, RepVGG~\cite{ding2021repvgg} proposes an extended $3\times3$ \textit{Conv} including $1\times1$ \textit{Conv} and \textit{residual connection}; DBB~\cite{ding2021diverse} proposes a diverse branch block to replace the original $K\times K$ \textit{Conv}, and each branch in the block can be transformed to a $K\times K$ \textit{Conv}; RepNAS~\cite{zhang2021repnas} aims to search DBB branches using neural architecture search (NAS).  Table~\ref{tab:operation_space} 
summarizes the detailed operation spaces. 

We end this section by showing a common caveat of Rep and its variants. Concretely, current methods simply re-parameterize all the candidate branches into an augmented network at the beginning of training, leading to a significant increase in memory and time consumption. For example, DBB has $\sim 2.3\times$ FLOPs and parameters on ResNet-18, which costs $\sim 1.7\times$ GPU days to train on ImageNet. Moreover, existing Rep and its variants generally involve redundant operations, which may introduce noise to the outputs and degrade the learning performance. A naive approach to address the above issues is training the augmented network with the effective operations. However, since re-parameterization can be nested, the oracle effective operations may be exhaustive to determine. With this regard, the desire of this study is developing an effective algorithm that utilizes the training information to incrementally find suitable structures.

\section{Dynamic Re-parameterization (DyRep)}

Here we propose Dynamic re-parameterization (DyRep) to seek the optimal structures with minimal cost by conservatively re-parameterizing the original network. We achieve the dynamic structure adaptation by extending the re-parameterization techniques in training. Supported by Rep that  transforms structures without changing  outputs, DyRep can evolve the structures flexibly during training and convert them back to the original network in inference. 

\subsection{Minimizing Loss with Dynamic Structures} \label{sec:expansion}

We follow Figure.~\ref{fig:figure2} to elaborate on the algorithmic implementation of DyRep.
In the training process, DyRep focuses on those operations in the   augmented network that have more contributions to the decrease of loss.
Consequently, instead of naively equipping all  operations with diverse branches adopted in DBB,
DyRep re-parameterizes  the operations that contribute the most to the loss. 

In DyRep, the contributions of  different operations towards decreasing the loss are measured by the gradient information. That is, an operation with small gradients contributes less to the decrease of loss, and is hence more likely to be redundant. Notably, a similar idea has also been broadly used in network pruning  \cite{lee2018snip,wang2019picking,tanaka2020pruning}, which endows an importance score for weights. Nevertheless, in contrast with network pruning that concentrates on the redundant (unimportant) weights, DyRep is more interested in identifying those  operations whose weights correspond to larger gradients. 

We next explain the score metric adopted in DyRep to identify the operations with  high contributions to decrease the loss. Many gradient-based score metrics~\cite{lee2018snip,wang2019picking,tanaka2020pruning} have been proposed to measure the saliency of weights. A recent study~\cite{tanaka2020pruning} proposes a score metric \textit{synflow} to avoid layer collapse when performing parameter pruning, \ie,
\begin{equation}
    \mathcal{S}_p(\theta_i) = \frac{\partial \L}{\partial \theta_i} \odot \theta_i ,
\end{equation}
where $\L$ is the loss function of a neural network with parameters $\bm{\theta}$, $\mathcal{S}_p$ is per-parameter saliency and $\theta_i\in\bm{\theta}$, and $\odot$ denotes Hadamard product. We extend $\mathcal{S}_p$ to score an entire operation by summing over all its parameters, i.e,
\begin{equation} \label{eq:saliency}
    \mathcal{S}_o(\bm{\theta}^{(i)}) = \sum_j^{n} \mathcal{S}_p(\theta^{(i)}_j) ,
\end{equation}
where $\bm{\theta}^{(i)}$ denotes the parameters in operation $o^{(i)}$. By leveraging Eq.(\ref{eq:saliency}), we gradually re-parameterize the operation with the largest $\mathcal{S}_o$ \wrt the accumulated training loss in every $t$ epochs, as described in Algorithm~\ref{alg:train_with_dyrep}. Note that DyRep works on all  $K\times K$ convolutions, including the newly-added Rep convolutions. This implies that our method can re-parameterize the operations recursively for even richer forms.

To re-parameterize those identified operations dynamically, we further extend the Rep technique to transform a single learned convolution into a DyRep block with random initialized weights during training. Our DyRep block consists of diverse Rep branches and computes the input feature to accumulate all its branches. The DyRep is equipped with all the candidate operations in Table~\ref{tab:operation_space}. Its weights are initialized with 
the proposed training-aware re-parameterization rule detailed below, where its illustration is shown in the left panel of Figure~\ref{fig:figure3}.

\textbf{Training-aware re-parameterization.}\ \ Suppose we have a set of randomly initialized operations $\{ o^{(1)}, ..., o^{(n)} \}$ to be added to the original operation $o^{ori}$ with weights $\bm{F}^{(ori)}$, then the new output of the expanded block becomes 
\begin{align}
    \begin{split}
        \bm{O'} &= \bm{I} \circledast \bm{F}^{(ori)} + \bm{I} \circledast \bm{F}^{(1)} + \cdots + \bm{I} \circledast \bm{F}^{(n)} .
    \end{split}
\end{align}
The output of the original operation $\bm{O} = \bm{I} \circledast \bm{F}^{(ori)}$ would be varied by new additional features. For a consistent output, we adopt Rep to transform the original weights $\bm{F}^{(ori)}$ using Eq.(\ref{eq:add}). The new weights $\bm{F}^{(ori')}$ yields
\begin{equation} \label{eq:grow}
    \bm{F}^{(ori')} \leftarrow \bm{F}^{(ori)} - (\bm{F}^{(1)} + \cdots + \bm{F}^{(n)}) ,
\end{equation}
where $\bm{F}^{(n)}$ is the transformed weights of operation $o^{(n)}$.

\textbf{Stabilizing training with batch normalization.}\ \ 
In Eq.(\ref{eq:grow}), if we initialize the added operations with large weights, the weights of the original operation will change a lot and thus its training would be disturbed. Fortunately, all our branches follow a \textit{OP-BN} paradigm, in which the last batch normalization (BN) layer~\cite{ioffe2015batch}
computes the input $\bm{x}$ as
\begin{equation}
    \mbox{BN}(\bm{x};\bm{\gamma},\bm{\beta}) = \bm{\gamma}\frac{\bm{x} - \mbox{E}(\bm{x})}{\sqrt{\mbox{Var}(\bm{x})}} + \bm{\beta} ,
\end{equation}
where $\bm{\gamma}$ and $\bm{\beta}$ are learnable weights for scaling and shifting the normalized value.
If we set $\bm{\gamma}$ to a small value and make $\bm{\beta}=0$, the transformed weights $\bm{F}$ of this branch will be small and the addition of branches will have a minor impact to the original weights. As a result, the functionality (representational ability) of the original convolution would be retained. We set $\bm{\gamma}=0.01$ in our experiments.

Besides, transforming weights of a branch into the weights of a single convolution needs to fuse the BN layer into convolution, which constructs new weight $\bm{F}'$ and bias $\bm{b}$ for every channel $j$ as
\begin{equation}
    \bm{F}_j' \leftarrow \frac{\gamma_j}{\sigma_j}\bm{F}_j,\quad b_j' \leftarrow \frac{(b_j-\mu_j)\gamma_j}{\sigma_j}+\beta_j ,
\end{equation}
where $\bm{\mu}$ and $\bm{\sigma}$ denote the accumulated running mean and variance in BN, respectively. For a randomly initialized branch, its $\bm{\mu}$ and $\bm{\sigma}$ in BN are initialized with $0$ and $1$, thus directly fusing the initialized BN into convolution would result in inaccurate weights. For an exact re-parameterization, we use $20$ batches of training data to calibrate the BN statistics before changing the weights in Eq.(\ref{eq:grow}). Note that the cost of BN calibration could be negligible since it does not require gradient computation.

\begin{algorithm}[h]
	\caption{Training with DyRep.}
	\label{alg:train_with_dyrep}
	
	\begin{algorithmic}[1]
	    \REQUIRE{Original model $\M$, total training epochs $E_\mathrm{tr}$, total iterations $N_\mathrm{iter}$ in each epoch, train dataset $\D_{tr}$, model update interval $t$, de-parameterization threshold $\lambda$.}
	    \FOR{$e=1,...,E_\mathrm{tr}$}
	        \FOR{$i=1,...,N_\mathrm{iter}$} 
	            \STATE train($\M$, $\D_{tr}$); \hfill$\triangleright$ \gray{train model for one iteration;}
	            \STATE $\{\mathcal{S}^{(i)}_{o}|o\in\M\} = \mbox{SCORE}(\M)$; \hfill$\triangleright$ \gray{scoring operations according to Eq.(\ref{eq:saliency});}
	        \ENDFOR
	        \IF{$e$ \% $t$ = 0}
	            \STATE $\mathcal{S}_{o} =  \frac{1}{N_\mathrm{iter}}\sum_{i=1}^{N_\mathrm{iter}}\mathcal{S}^{(i)}_{o}, \forall o \in \M$ \hfill$\triangleright$ \gray{average over all iterations;}
	            \STATE $o^* = \argmax_{o}\{\mathcal{S}_o|o\in\M\}$; \hfill$\triangleright$ \gray{locate the most important operation;}
	            \STATE $\M = \mbox{REP}(\M, o^*)$; \hfill$\triangleright$ \gray{re-parameterize operation to a DyRep block according to Section~\ref{sec:expansion}};
	            \STATE $\M = \mbox{DEP}(\M, \lambda)$; \hfill$\triangleright$ \gray{de-parameterize branches according to Section~\ref{sec:dep};}
	        \ENDIF
	    \ENDFOR
	    \STATE deploy model $\M$ back to the original model;
	    \RETURN inference model $\M$.
	\end{algorithmic}
\end{algorithm}
\subsection{De-parameterizing for Better Efficiency} \label{sec:dep}

During training, we dynamically locate the most important operation using Eq.(\ref{eq:saliency}) and re-parameterize it to a DyRep block; as Rep may introduce redundant operations, we also design a rule to endow the capability of discarding operations, which we call \textit{de-parameterization} (Dep). In the sequel, we will discuss how the Dep works. 

As in Section~\ref{sec:expansion}, we set the $\bm{\gamma}$ of BN to a small value to stabilize training; considering that BN first normalizes the input $\bm{x}$ with the same magnitude, then uses $\bm{\gamma}$ and $\bm{\beta}$ to scale and shift the values, which means that the $\bm{\gamma}$ and $\bm{\beta}$ control the magnitudes of branches' outputs. If a branch has an obviously smaller output values than other branches, we may think it makes a minor contribution to the final outputs. Note that setting $\gamma = 0$ effectively zeros out the output.

In this way, we now propose an approach to find the redundant operations by comparing the scale factors of BN layers as ~\cite{liu2017learning,gordon2018morphnet,ye2018rethinking,huang2018data}. Concretely, we use the L1 norm of $\bm{\gamma}$ of the last BN layer to represent the importance $s_j$ of branch $j$, \ie, 
\begin{equation}
    s_j = \frac{1}{C}\sum_{k=1}^{C}|\gamma_k| ,
\end{equation}
where $C$ denotes the number of channels in BN. Since the BN layers are initialized with the same weights at the beginning, we can thus take the branches with \textit{significantly low} weights as the one to cut after a period of training. In our method, we simply select the one with $s_j < \mbox{Mean}(\{s_j\}_{j=1}^{n})$ when branches evolve to be sufficiently distinguishable satisfying $\sqrt{\mbox{Var}(\{s_j\}_{j=1}^{n})}>\lambda$, where $\lambda$ is a threshold and $\lambda = 0.02$ would suffice empirically. 

\textbf{De-parameterization.}\ \ Similar to the training-aware Rep technique, for a DyRep block with operations $\{ o^{(ori)}, o^{(1)}, ..., o^{j}, ..., o^{n} \}$, if we want to remove operation $o^{j}$ yet make the outputs consistent, we absorb the weights of $o^{j}$ into $o^{(ori)}$, \ie,
\begin{equation} \label{eq:cut}
    \bm{F}^{(ori')} \leftarrow \bm{F}^{(ori)} + \bm{F}^{(j)} ,
\end{equation}
then we can safely cut the operation $o^{j}$. Since the redundant operation has a small scale factor $\bm{\gamma}$, its removal would also have a minor influence on the weights of original convolution, and the training stability  would be retained.

\subsection{Progressive Training with Rep and Dep}

With the proposed re-parameterization (Rep) and de-parameterization (Dep) techniques, we can dynamically enrich the desired operations and discard some redundant operations simultaneously. Combining both Rep and Dep, the networks can be enhanced with great efficiency. The overall training procedure is summarized in Algorithm~\ref{alg:train_with_dyrep}. More precisely, DyRep repeatedly proceeds Rep and Dep in every $t$ epochs, as shown in Lines 3-7. The Rep procedure consists of three components, i.e., selecting the operation with the largest saliency score $S_o$, expanding it with randomly initialized operations, and  modifying the weights of the original operation according to Eq.(\ref{eq:grow}) to ensure that the expanded block has the same output as the original operation. The Dep procedure also contains three parts, i.e., finding redundant operations using $\bm{\gamma}$ of BN for each branch, removing those redundant operations, and  modifying the weights of original operation according to Eq.(\ref{eq:cut}).

\section{Experiments}
\subsection{Training Strategies}
\textbf{CIFAR.} Following DBB~\cite{ding2021diverse}, we train VGG-16 models with batch size $128$, a cosine learning rate which decays $600$ epochs is adopted with initial value $0.1$, and use SGD optimizer with momentum 0.9 and weight decay $1\times 10^{-4}$. We set structure update interval $t=15$ in our method.

\textbf{ImageNet.} In Table~\ref{tab:exp_imgnet_dbb}, we train models with the same strategies as DBB~\cite{ding2021diverse}. Concretely, we train ResNet-18 and ResNet-50 for $120$ epochs with a total batch size $256$ and color jitter data augmentation, a cosine learning rate strategy with initial value $0.1$ is adopted, and the optimizer is SGD with momentum 0.9 and weight decay $1\times 10^{-4}$. For MobileNet, we train the model with weight decay $4\times10^{-5}$ for $90$ epochs. While for VGG-style models, following RepVGG~\cite{ding2021repvgg}, we train the models for $200$ epochs with a strong data augmentation (Autoaugment~\cite{cubuk2019autoaugment} and label smooting), except for \textit{DyRep-A2} we use a simple data augmentation and train it for $120$ epochs. In our method, we set structure update interval $t=5$ for $120$-epoch and $200$-epoch training; for $300$ epoch training, we set $t=10$.

\begin{table}
	\renewcommand\arraystretch{1.17}
	\setlength\tabcolsep{0.6mm}
	\centering
	\caption{Results of base model VGG-16~\cite{simonyan2014very} on CIFAR datasets. Results are reported based on our implementation with the same training strategies. The FLOPs and parameters in the table are average values in training. Training cost is tested on a NVIDIA Tesla V100 GPU.}
	\vspace{-2mm}
	\label{tab:exp_cifar}
	\footnotesize
	\begin{tabular}{llcccc}
		\Xhline{2\arrayrulewidth}
		\multirow{2}*{Dataset} & Rep & Cost & Avg. FLOPs & Avg. params & ACC \\
		~ & method & (GPU hrs.) & (M) & (M) & (\%)\\
        \hline
        \multirow{3}*{CIFAR-10} & Origin & 2.4 & 313 & 15.0 & 94.68$\pm$0.08 \\
        ~ & DBB & 9.4 & 728 & 34.7 & 94.97$\pm$0.06\\
        ~ & DyRep & 6.9 & 597 & 26.4 & \textbf{95.22}$\pm$0.13\\
        \hline
        \multirow{3}*{CIFAR-100} & Origin & 2.4 & 313 & 15.0 & 73.69$\pm$0.12 \\
        ~ & DBB & 9.4 & 728 & 34.7 & 74.04$\pm$0.08\\
        ~ & DyRep & 6.7 & 582 & 27.1 & \textbf{74.37}$\pm$0.11\\
		\Xhline{2\arrayrulewidth}
	\end{tabular}
\end{table}
\begin{table}
	\renewcommand\arraystretch{1.17}
	\setlength\tabcolsep{1.2mm}
	\centering
	\caption{Results of MobileNet~\cite{howard2017mobilenets} and ResNet~\cite{he2016deep} models on ImageNet dataset compared to DBB~\cite{ding2021diverse}. Training cost is tested on $8$ NVIDIA Tesla V100 GPUs. *: Our implementation.}
	\vspace{-2mm}
	\label{tab:exp_imgnet_dbb}
	\footnotesize
	\begin{tabular}{llcccc}
		\Xhline{2\arrayrulewidth}
		\multirow{2}*{Model} & Rep & Cost & Avg. FLOPs & Avg. params & ACC \\
		~ & method & (GPU days) & (G) & (M) & (\%)\\
        \hline
        \multirow{3}*{MobileNet} & Origin & 2.3 & 0.57 & 4.2 & 71.89 \\
        ~ & DBB & 4.2 & 0.61 & 4.3 & 72.88 \\
        ~ & DyRep & 2.4 & 0.58 & 4.3 & \textbf{72.96} \\
        \hline
        \multirow{3}*{ResNet-18} & Origin & 4.8 & 1.81 & 11.7 & 69.54\\
        ~ & DBB & 8.1 & 4.13 & 26.3 & 70.99\\
        ~ & DyRep & 6.3 & 2.42 & 16.9 & \textbf{71.58}\\
        \hline
        \multirow{3}*{ResNet-34} & Origin & 5.3 & 3.66 & 21.8 & 73.31\\
        ~ & DBB$^*$ & 12.8 & 8.44 & 49.9 & 74.33\\
        ~ & DyRep & 7.7 & 4.72 & 33.1 & \textbf{74.68}\\
        \hline
        \multirow{3}*{ResNet-50} & Origin & 7.5 & 4.09 & 25.6 & 76.14\\
        ~ & DBB & 13.7 & 6.79 & 40.7 & 76.71\\
        ~ & DyRep & 8.5 & 5.05 & 31.5 & \textbf{77.08}\\
		\Xhline{2\arrayrulewidth}
	\end{tabular}
	\vspace{-4mm}
\end{table}

\subsection{Compare with DBB}

We first compare our method with the baseline method DBB~\cite{ding2021diverse} on CIFAR and ImageNet datasets. For fair comparisons, we conduct experiments using the same models and training strategies as DBB, and the results are summarized in Table~\ref{tab:exp_cifar} and Table~\ref{tab:exp_imgnet_dbb}. The FLOPs and params in the tables are averaged values in training. The results on both CIFAR and ImageNet datasets show that, our method enjoys significant performance improvements compared to the original and DBB models. Meanwhile, the training cost of our DyRep is obviously lower than DBB. For example, DBB costs $13.7$ GPU days to train ResNet-50 on ImageNet, while our DyRep costs $8.5$ GPU days ($\sim 38\%$ less) and gets $0.37\%$ improvement on accuracy.

\subsection{Better Performance with RepVGG}
RepVGG~\cite{ding2021repvgg} proposes a series of VGG-style networks and achieves competitive performance to current state-of-the-art models. We adopt our DyRep on these VGG networks for better performance on ImageNet. Results are summarized in Table~\ref{tab:imagenet}. Our DyRep obtains higher accuracies with the same transformed models compared to RepVGG since it only adopts two branches $1\times1$ \textit{Conv} and \textit{residual connection}. Meanwhile, compared to the results (ODDB) obtained by RepNAS~\cite{zhang2021repnas}, our method also achieves higher performance. For example, our \textit{DyRep-B3} achieves $81.12\%$ accuracy, outperforms RepVGG-B3 and ODBB(B3) by $0.6\%$ and $0.15\%$, respectively.

\begin{table}
	\renewcommand\arraystretch{1.17}
	\setlength\tabcolsep{2mm}
	\centering
	\caption{Results on ImageNet. The FLOPs and parameters are measured on inference models. Training cost is tested on 8 NVIDIA Tesla V100 GPUs. The baseline ACCs and speeds are reported by RepVGG~\cite{ding2021repvgg}.}
	\vspace{-2mm}
	\label{tab:imagenet}
	\footnotesize
	\begin{tabular}{lcccc}
		\Xhline{2\arrayrulewidth}
		\multirow{2}*{Model} & Speed & FLOPs & params & ACC \\
		~ & (examples/second) & (G) & (M) & (\%)\\
        \hline
        ResNet-34 & 1419 & 3.7 & 21.78 & 74.17\\
        RepVGG-A2 & 1322 & 5.1 & 25.49 & 76.48 \\
        ODBB(A2) & 1322 & 5.1 & 25.49 & 76.86 \\
        \textbf{DyRep-A2} & 1322 & 5.1 & 25.49 & 76.91\\
        ResNet-50 & 719 & 3.9 & 25.53 & 76.31\\
        \hline
        RepVGG-B2g4 & 581 & 11.3 & 55.77 & 79.38  \\
        \textbf{DyRep-B2g4} & 581 & 11.3 & 55.77 & 80.12\\
        ResNeXt-50 & 484 & 4.2 & 24.99 & 77.46 \\
        ResNet-101 & 430 & 7.6 & 44.49 & 77.21\\
        \hline
        RepVGG-B3 & 363 & 26.2 & 110.96 & 80.52 \\
        ODBB(B3) & 363 & 26.2 & 110.96 & 80.97 \\
        \textbf{DyRep-B3} & 363 & 26.2 & 110.96 & 81.12 \\
        ResNeXt-101 & 295 & 8.0 & 44.10 & 78.42\\
		\Xhline{2\arrayrulewidth}
	\end{tabular}
\end{table}

\subsection{Results on Downstream Tasks}
We transfer our ImageNet-pretrained ResNet-50 model to downstream tasks object detection and semantic segmentation to validate our generalization. Concretely, we use the pretrained model as the backbone of the downstream algorithms FPN~\cite{lin2017feature} and PSPNets~\cite{zhao2017pyramid} on COCO and Cityscapes datasets, respectively, then report their evaluation results on validation sets. Moreover, since our DyRep can evolve structures during training, we can directly load the plain weights of ResNet-50 and augment its structure in the training of downstream tasks, without training on ImageNet. Therefore, we conduct experiments to adopt DyRep in the training of downstream tasks. The results on Table~\ref{tab:downstream} show that, by directly transferring the trained model to downstream tasks (refers \textit{C} in the table), our DyRep can obtain better performance compared to original ResNet-50 and DBB. When we adopt DyRep to update structures in downstream tasks, the performance can be further improved.

\begin{table}
	\renewcommand\arraystretch{1.17}
	\setlength\tabcolsep{2mm}
	\centering
	\caption{Results on object detection and semantic segmentation tasks. Rep stage \textit{C+D} denotes the Rep method is adopted on both ImageNet training and downstream tasks.}
	\vspace{-2mm}
	\label{tab:downstream}
	\footnotesize
	\begin{tabular}{lccccc}
		\Xhline{2\arrayrulewidth}
		\multirow{2}*{Backbone} & Rep & Rep & ImageNet & COCO & Cityscapes\\
		~ & method & stage & top-1 & mAP & mIOU\\
		\hline
		ResNet-50 & - & - & 76.13 & 37.4 & 77.85 \\
		ResNet-50 & DBB & C & 76.78 & 37.8 & 78.18 \\
		\hline
		ResNet-50 & DyRep & C & 77.08 & 38.0 & 78.32 \\
		ResNet-50 & DyRep & D & 76.13 & 37.7 & 78.09 \\
		ResNet-50 & DyRep & C+D & 77.08 & \textbf{38.1} & \textbf{78.49} \\
		\Xhline{2\arrayrulewidth}
	\end{tabular}
\end{table}

\subsection{Ablation Studies}

\textbf{Effect of de-parameterization.} We propose de-parameterization (Dep) to discard those redundant Rep branches. Now we conduct experiments to validate its effectiveness. As summarized in Table~\ref{tab:dep}, combining Rep and Dep achieves the best performance, while using DyRep without Dep can also boost the performance, but the training efficiency and accuracy will drop.
\begin{table}[htbp]
	\renewcommand\arraystretch{1.17}
	\setlength\tabcolsep{3mm}
	\centering
	\caption{Results of DyRep on CIFAR-10 dataset w/ or w/o Dep.}
	\vspace{-2mm}
	\label{tab:dep}
	\footnotesize
	\begin{tabular}{lccc}
		\Xhline{2\arrayrulewidth}
		Method & FLOPs (M) & Params (M) & ACC (\%)\\
		\hline
		Origin & 313 & 15.0 & 94.68$\pm$0.08\\
		DyRep w/ Dep & 597 & 26.4 & \textbf{95.22}$\pm$0.13\\
		DyRep w/o Dep & 658 & 29.3 & 95.03$\pm$0.15\\
		\Xhline{2\arrayrulewidth}
	\end{tabular}
\end{table}

\begin{figure}[ht]
    \centering
    \includegraphics[width=0.7\linewidth]{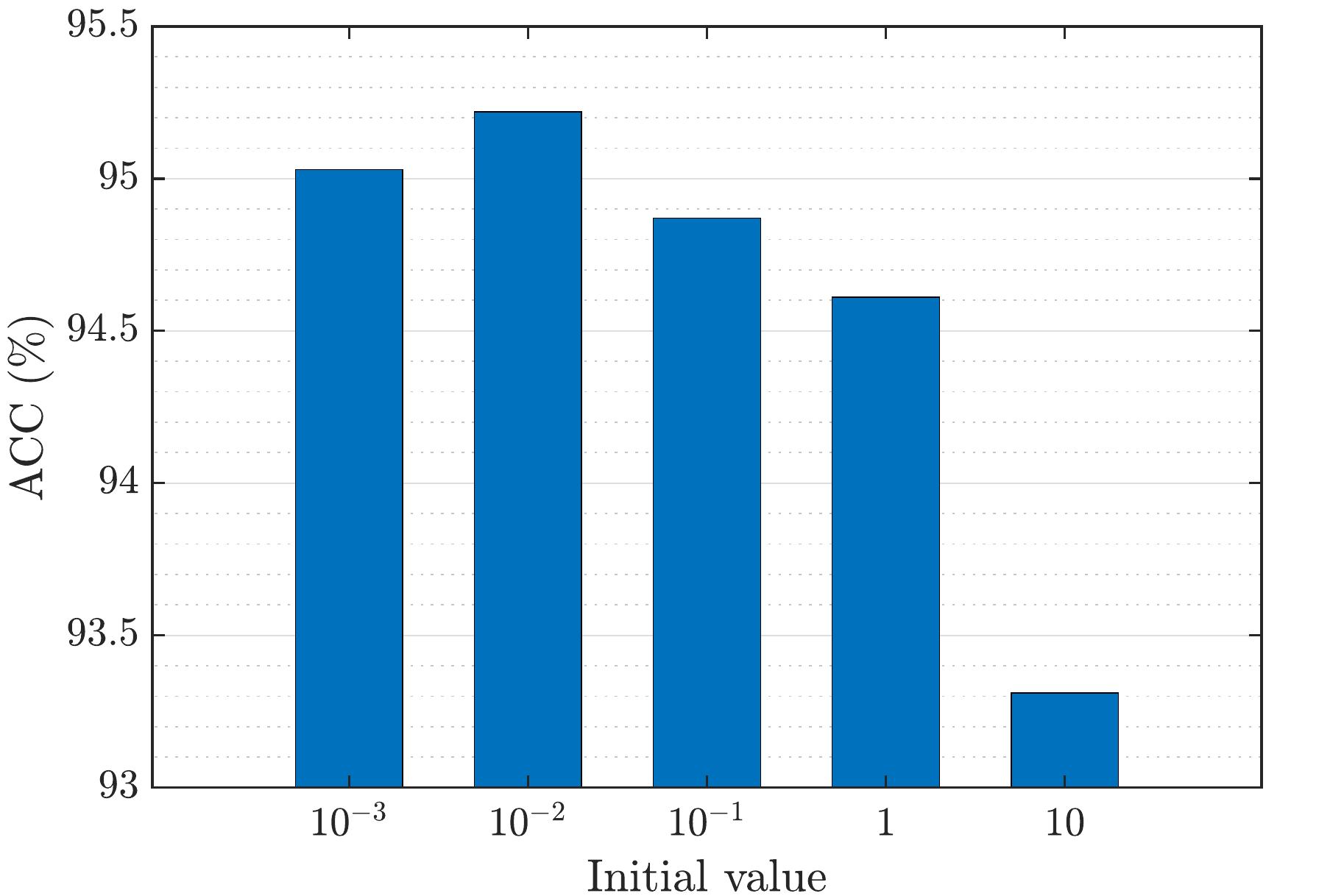}
    \vspace{-2mm}
    \caption{Evaluation results on CIFAR-10 with different initial values of $\bm{\gamma}$ in BN.}
    \label{fig:initial_gamma}
    \vspace{-4mm}
\end{figure}

\textbf{Effects of different initial scale factors.} To stabilize the training, we initialize the last BN layer in each newly-added branch with a small scale factor. Here we conduct experiments to show the effects of different initial values of scale factors. As shown in Figure~\ref{fig:initial_gamma}, the performance of different scale factors varies a lot. For a large initial value of $10$, the original weights would be changed dramatically; therefore, its accuracy is far below the others. While for a tiny initial value, the performance decreases since the added operations contribute trivially to the outputs. According to the results, we initialize $\gamma$ with $0.01$ in our main experiments.

\textbf{Retraining obtained structures from scratch.} To validate the effect of our dynamic structure adaptation, we retrain the intermediate structures obtained in DyRep training on CIFAR-10. As shown in Table~\ref{tab:retrain}, training with DyRep obtains better accuracy compared to training fixed intermediate structures. A possible reason is that the best structures are different at different training phases (epochs). Our DyRep can dynamically adapt the structures to boost the performance in the whole training, thus getting better performance, while the obtained structures at different epochs might not best suit the other epochs. 
\begin{table}[htbp]
	\renewcommand\arraystretch{1.17}
	\setlength\tabcolsep{3mm}
	\centering
	\caption{Retraining results of intermediate structures on CIFAR-10. We fix the structures and retrain them with the same strategy. DyRep denotes training with DyRep, and DyRep-$N$ denotes the structure of DyRep at $N$th epoch.}
	\vspace{-2mm}
	\label{tab:retrain}
	\footnotesize
	\begin{tabular}{lccc}
		\Xhline{2\arrayrulewidth}
		Model & FLOPs (M) & Params (M) & ACC (\%)\\
		\hline
		Origin & 313 & 15.0 & 94.68$\pm$0.08\\
		DyRep-200 & 466 & 18.2 & 94.83$\pm$0.06\\
		DyRep-400 & 587 & 20.3 & 95.04$\pm$0.07\\
		DyRep-600 & 766 & 24.4 & 94.93$\pm$0.05\\
		\hline
		DyRep & 597 & 26.4 & \textbf{95.22$\pm$0.13}\\
		\Xhline{2\arrayrulewidth}
	\end{tabular}
\end{table}

\begin{figure}[h]
    \centering
    \includegraphics[width=0.66\linewidth]{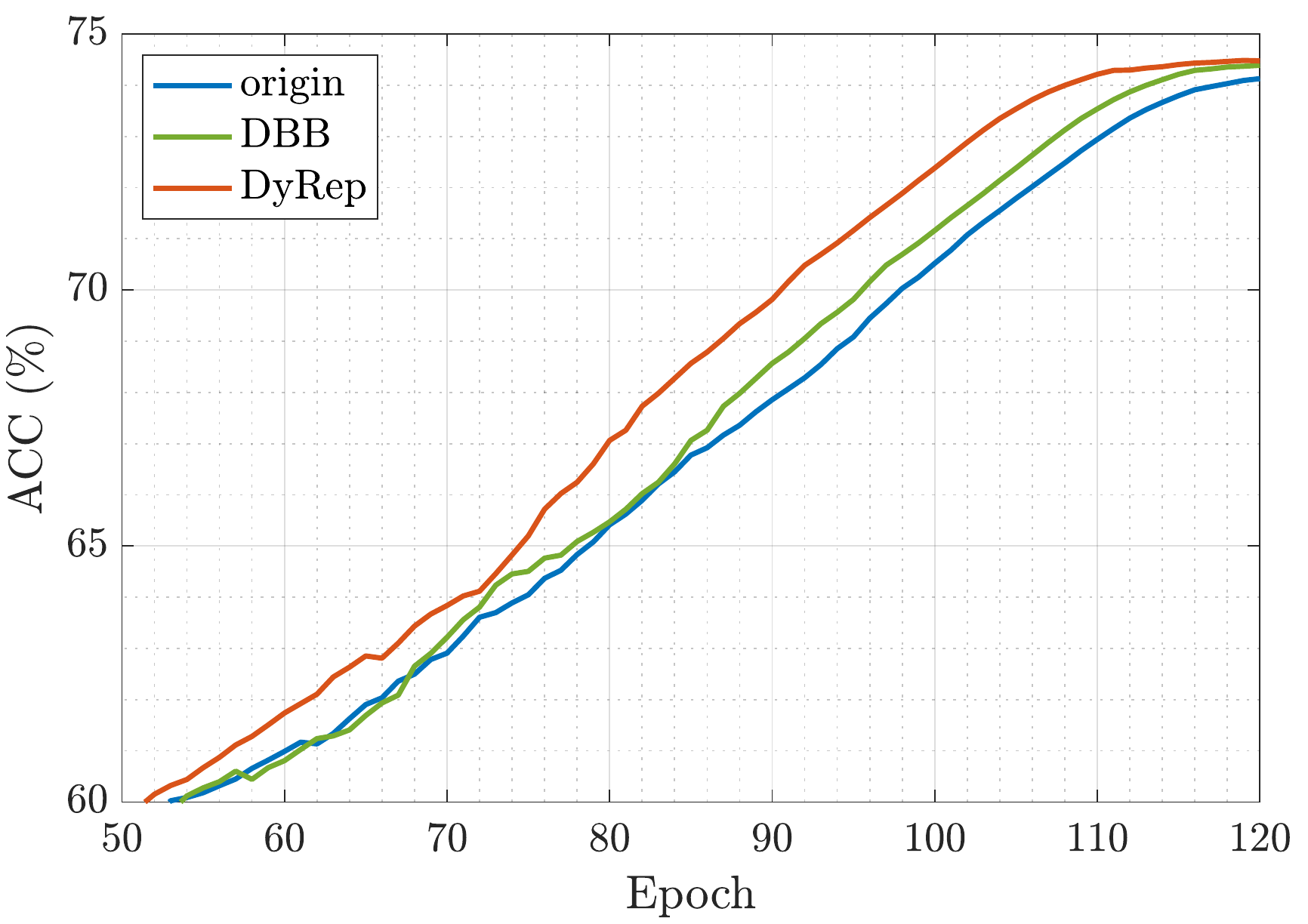}
    \vspace{-2mm}
    \caption{Evaluation curves (smoothed) of ResNet-34 models in training using DyRep, DBB, and original models.}
    \label{fig:res34_curves}
\end{figure}

\textbf{Visualization of convergence curves.} We visualize the convergence curves of our method with comparisons of the original ResNet-34 model and DBB. As shown in Figure~\ref{fig:res34_curves}, our DyRep enjoys better convergence during the whole training. It might be because our method efficiently involves effective operations without additional training overhead to the redundant operations.

\begin{figure}[h]
    \centering
    \includegraphics[width=0.9\linewidth]{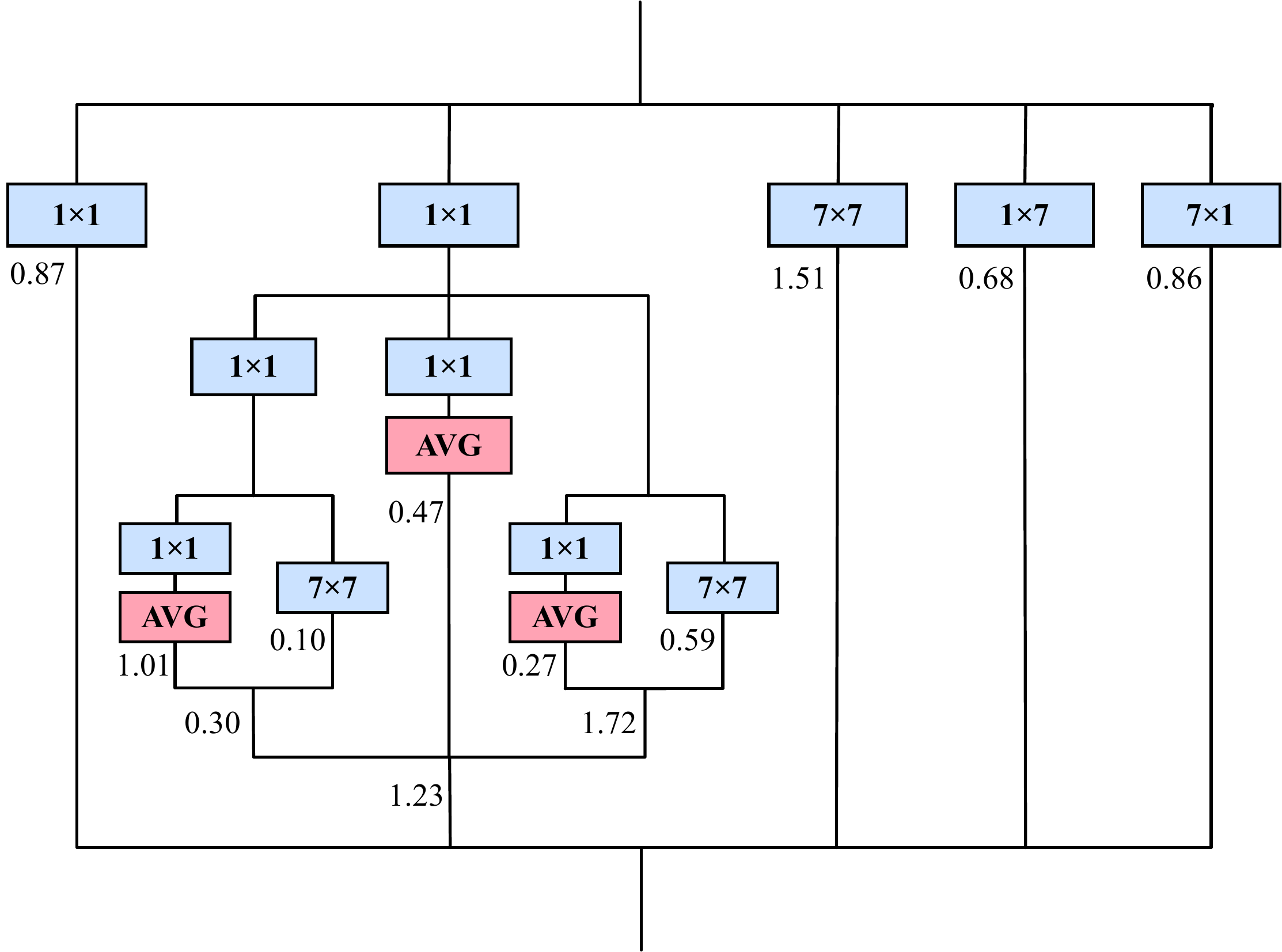}
    \vspace{-0mm}
    \caption{Visualization of the final augmented structures of the fist convolution layer in ResNet-18. The weight below each branch denotes its importance factor $s$.}
    \label{fig:vis_res18}
    \vspace{-4mm}
\end{figure}

\textbf{Visualization of augmented structures.}
We visualize the first convolution layer in our trained ResNet-18 network in Figure~\ref{fig:vis_res18}. We can see that the original $7\times7$ \textit{Conv} has the largest weight, and all the convolutions with kernel size $K > 1$ are equipped since they help extract information in the previous stages of the network. While on the $1\times1$-$7\times7$ branch, it expands recursively and leverages average pooling to enrich the features. 

\subsection{Limitations}
DyRep can efficiently enhance the performance of various CNNs, which could be treated as a universal mechanism to boost the training without changing the inference structures. However, its operation space is restricted by the requirements of equivalence in re-parameterization.

\section{Conclusion}
We propose DyRep, a novel method to bootstrap training with \textit{dynamic} re-parameterization (Rep). Concretely, to enhance the representational capacity of the operation, which contributes most to the performance, we extend the existing Rep technique to dynamically re-parameterize it and propose an initialization strategy of BN to stabilize the training. Moreover, with the help of the scale factors in BN layers, we propose a mechanism to discard those redundant operations. As a result, the training efficiency and performance can be further improved. Extensive experiments show that our DyRep enjoys significant efficiency and performance improvements compared to existing Rep methods. 

\section*{Acknowledgements}
This work was supported in part by the Australian Research Council under Project DP210101859 and the University of Sydney SOAR Prize.

{\small
\bibliographystyle{ieee_fullname}
\bibliography{main}
}

\newpage
\appendix
\onecolumn

\section{More Ablation Studies}

\subsection{Performance on different operation spaces}
Compared to the DBB~\cite{ding2021diverse} using $4$ branches to build the augmented network, DyRep further adds $3$ operations for more flexible structures. We now conduct experiments to compare our DyRep with DBB using the same operation spaces for fair comparisons. As summarized in Table~\ref{tab:ab_operation_space}, we train ResNet-50 using DyRep and DBB with $4$ operations and $7$ operations on ImageNet and report their validation accuracies. When using the same operation space as DBB, our DyRep can also enjoy significant efficiency and performance improvements. Besides, if we adopt DBB on our larger operation space, its training cost will be higher, and our superiority will be more significant. Comparing $4$ branches and $7$ branches, the larger one achieves better accuracy because of more diverse representations.

\begin{table}[h]
	\renewcommand\arraystretch{1.3}
	\setlength\tabcolsep{3mm}
	\centering
	\caption{Evaluation results of ResNet-50 on ImageNet dataset using different numbers of Rep operations. *: our implementation.}
	\vspace{-2mm}
	\label{tab:ab_operation_space}
	\footnotesize
	\begin{tabular}{ccllll}
		\Xhline{2\arrayrulewidth}
        Rep method & \#operations & FLOPs & Params & Training cost & ACC (\%)\\
        \hline
        Origin & 1 & 4.09 & 25.6 & 7.5 & 76.14\\
        \hline
        DBB & 4 & 6.79 & 40.7 & 13.7 & 76.71\\
        DyRep & 4 & \textbf{4.93} (\green{-27.4\%})& \textbf{30.2} (\green{-25.8\%}) & \textbf{8.1} (\green{-40.9\%}) & \textbf{76.98} (\green{+0.27\%})\\
        \hline
        DBB$^*$ & 7 & 8.02 & 48.3 & 17.3 & 76.87\\
        DyRep & 7 & \textbf{5.05} (\green{-37.0\%}) & \textbf{31.5} (\green{-34.8\%}) & \textbf{8.5} (\green{-50.9\%}) & \textbf{77.08} (\green{+0.21\%})\\
        
		\Xhline{2\arrayrulewidth}
	\end{tabular}
	\vspace{-4mm}
\end{table}

\subsection{Effects of different update intervals $t$ of DyRep}
We update the structures of the network in every $t$ epoch. If the structures are expanded more frequently, the final network will be larger. As summarized in Table~\ref{tab:interval}, we train the models with different update intervals $t$, and the results show that for a small $t$, the accuracy will be further improved, but the training cost also increases accordingly. For efficiency consideration, we choose a moderate frequency $t=15$ on CIFAR-10.

\begin{table}[h]
	\renewcommand\arraystretch{1.3}
	\setlength\tabcolsep{4mm}
	\centering
	\caption{Evaluation results of VGG-16 on CIFAR-10 with different update interval $t$.}
	\vspace{-2mm}
	\label{tab:interval}
	\footnotesize
	\begin{tabular}{cccccc}
		\Xhline{2\arrayrulewidth}
		Rep method & Update interval $t$ & Cost (GPU hours) & FLOPs (M) & Params (M) & ACC (\%)\\
		\hline
		origin & - & 2.4 & 313 & 15.0 & 94.68\\
		DBB & - & 9.4 & 728 & 34.7 & 94.97\\
		\hline
		DyRep & 5 & 13.3 & 1575 & 83.4 & 95.39\\
		DyRep & 10 & 8.8 & 992 & 33.6 & 95.33 \\
		DyRep & 15 & 6.9 & 597 & 26.4 & 95.22 \\
		DyRep & 30 & 5.8 & 522 & 23.7 & 94.91\\
		DyRep & 50 & 4.1 & 430 & 20.3 & 94.82\\
		\Xhline{2\arrayrulewidth}
	\end{tabular}
	\vspace{-4mm}
\end{table}

\subsection{Effects of different scoring metrics in our Rep}
Many metrics~\cite{lee2018snip,wang2019picking,tanaka2020pruning} are proposed to measure the saliency score of weights in network pruning. In our paper, to choose the most suitable metric, we conduct experiments to evaluate these scoring metrics in DyRep. The experimented scoring metrics are summarized as follows.

For one operation with weights $\bm{\theta}$, its saliency score can be represented as following metrics:

\begin{itemize}
    \item \textit{random}: the score of each operation is generated randomly.
    \begin{equation}
        \mathcal{S}_o(\bm{\theta}) \stackrel{\mathrm{iid}}{\sim} \mathbb{R}^1 .
    \end{equation}
    \item \textit{grad\_norm}: A simple baseline of summing the Euclidean norm of the gradients.
    \begin{equation}
        \mathcal{S}_o(\bm{\theta}) = \left\|\frac{\partial\L}{\partial\bm{\theta}}\right\|_2 .
    \end{equation}
    \item \textit{snip}~\cite{lee2018snip}:
    \begin{equation}
        \mathcal{S}_o(\bm{\theta}) = \sum_i^n\left|\frac{\partial\L}{\partial\theta_i}\odot\theta_i\right| .
    \end{equation}
    \item \textit{grasp}: \cite{wang2019picking} aims to improve \textit{snip} by approximating the change in gradient norm (instead of loss), therefore its \textit{grasp} metric is computed as
    \begin{equation}
        \mathcal{S}_o(\bm{\theta}) = \sum_i^n-(H\frac{\partial\L}{\partial\theta_i}\odot\theta_i) ,
    \end{equation}
    where $H$ denotes the Hessian matrix.
    \item \textit{synflow}: SynFlow~\cite{tanaka2020pruning} proposes a modified version (\textit{synflow}) to avoid layer collapse when performing parameter pruning.
    \begin{equation}
        \mathcal{S}_o(\bm{\theta}) = \sum_i^n\frac{\partial\L}{\partial\theta_i}\odot\theta_i .
    \end{equation}
    \item \textit{vote}: Inspired by \cite{abdelfattah2020zero}, 
    which leverages the above metrics to vote the decisions, we also provide a result of picking operations with the most votes.
\end{itemize}

We measure all the metrics above on CIFAR-10 dataset, as summarized in Table~\ref{tab:metrics}. The random baseline even worsens the performance as it could introduce some unnecessary disturbances to the original weights, showing that it is important in choosing operations. Besides, \textit{synflow} achieves the best performance compared to other metrics, and we thus adopt it as the scoring metric in our DyRep.

\begin{table}[h]
	\renewcommand\arraystretch{1.3}
	\setlength\tabcolsep{4mm}
	\centering
	\caption{Accuracies of VGG-16 using different scoring metrics in DyRep. \textit{Origin} denotes the original results of model without Rep.}
	\vspace{-2mm}
	\label{tab:metrics}
	\footnotesize
	\begin{tabular}{l|cc|cccc|c}
		\Xhline{2\arrayrulewidth}
        Dataset & origin & random & grad\_norm & snip & grasp & synflow & vote \\
        \hline
        CIFAR-10 & 94.68 & 94.25 & 94.71 & 94.97 & 94.82 & \textbf{95.22} & 95.03\\
        CIFAR-100 & 74.10 & 74.03 & 74.19 & 74.56 & 74.73 & \textbf{74.91} & 74.79\\
		\Xhline{2\arrayrulewidth}
	\end{tabular}
	\vspace{-4mm}
\end{table}

\subsection{Effects of training with DyRep for different epochs}
\begin{figure}[h]
    \centering
    \includegraphics[width=0.8\linewidth]{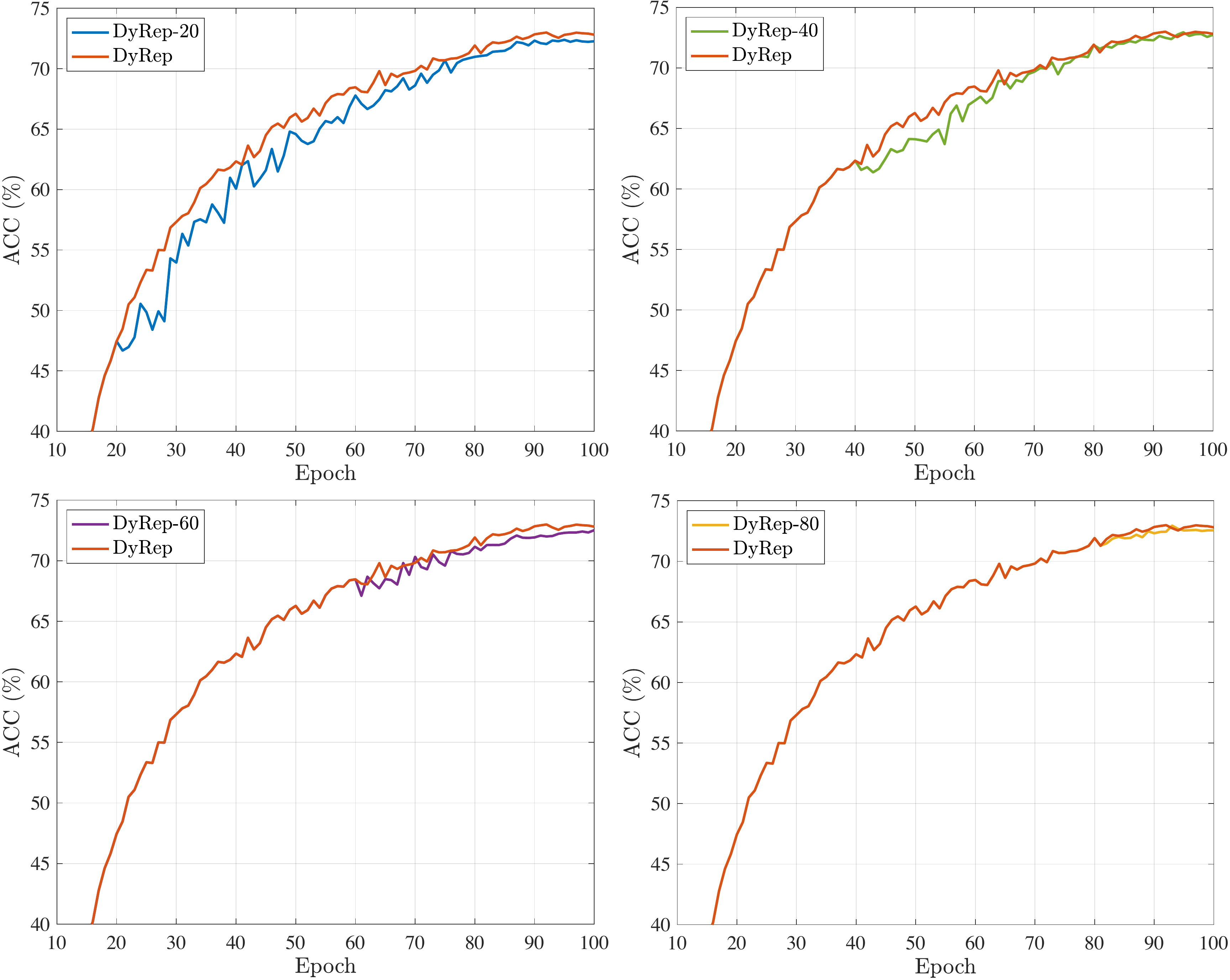}
    \caption{Training curves of adopting DyRep for different epochs on CIFAR-100. DyRep-$N$ denotes training with DyRep for the first $N$ epochs then fixing the structure for the latter $100-N$ epochs. DyRep with red line means adopting DyRep in the whole training.}
    \label{fig:dyrep_stage_curves}
    \vspace{-4mm}
\end{figure}

To validate the effectiveness of DyRep in boosting training, we adopt DyRep for the first $20$,$40$,$60$,$80$, and $100$ epochs, then train the remained epochs with fixed structures. As illustrated in Figure~\ref{fig:dyrep_stage_curves}, our DyRep can dynamically adapt the structures thus is more steady compared to training with fixed structures. Besides, adopting Rep can always obtain higher accuracy than fixed structures, showing that DyRep can improve performance in the whole training process.

\end{document}